\documentclass[iicol, pdflatex,sn-mathphys-num]{sn-jnl}

\usepackage{booktabs}
\usepackage{graphicx}%
\usepackage{multirow}%
\usepackage{amsmath,amssymb,amsfonts}%
\usepackage{amsthm}%
\usepackage{mathrsfs}%
\usepackage[title]{appendix}%
\usepackage{xcolor}%
\usepackage{textcomp}%
\usepackage{manyfoot}%
\usepackage{booktabs}%
\usepackage{algorithm}%
\usepackage{algorithmicx}%
\usepackage{algpseudocode}%
\usepackage{listings}%

\usepackage{multirow}
\usepackage{subcaption}
\usepackage{wrapfig}
\usepackage{adjustbox}

\usepackage{pifont}

\theoremstyle{thmstyleone}%
\theoremstyle{thmstyletwo}%

\theoremstyle{thmstylethree}%

\begin{document}

\title[Article Title]{StarryGazer: Leveraging Monocular Depth Estimation Models for Domain-Agnostic Single Depth Image Completion}


\author[1]{\fnm{Sangmin} \sur{Hong}}\email{mchiash2@snu.ac.kr}
\equalcont{These authors contributed equally to this work.}
\author[2]{\fnm{Suyoung} \sur{Lee}}\email{esw0116@snu.ac.kr}
\equalcont{These authors contributed equally to this work.}

\author*[1,2]{\fnm{Kyoung Mu} \sur{Lee}}\email{kyoungmu@snu.ac.kr}

\affil[1]{IPAI, Seoul National University, Seoul 08826, South Korea}
\affil[2]{Department of Electrical and Computer Engineering, ASRI, Seoul National University, Seoul 08826, South Korea}


\abstract{

The problem of depth completion involves predicting a dense depth image from a single sparse depth map and an RGB image.
Unsupervised depth completion methods have been proposed for various datasets where ground truth depth data is unavailable and supervised methods cannot be applied.
However, these models require auxiliary data to estimate depth values, which is far from real scenarios.
Monocular depth estimation (MDE) models can produce a plausible relative depth map from a single image, but there is no work to properly combine the sparse depth map with MDE for depth completion; a simple affine transformation to the depth map will yield a high error since MDE are inaccurate at estimating depth difference between objects.
We introduce \textbf{StarryGazer}, a domain-agnostic framework that predicts dense depth images from a single sparse depth image and an RGB image without relying on ground-truth depth by leveraging the power of large MDE models.
First, we employ a pre-trained MDE model to produce relative depth images. These images are segmented and randomly rescaled to form synthetic pairs for dense pseudo-ground truth and corresponding sparse depths.
A refinement network is trained with the synthetic pairs, incorporating the relative depth maps and RGB images to improve the model's accuracy and robustness.
StarryGazer shows superior results over existing unsupervised methods and transformed MDE results on various datasets, demonstrating that our framework exploits the power of MDE models while appropriately fixing errors using sparse depth information.
}




\maketitle

\section{Introduction}
\label{Introduction}
Navigating through an environment with incomplete sensory information can be likened to finding one's way in a dense fog with just a flashlight; While a general sense of the surroundings is possible, obtaining a complete visual picture remains elusive. Similarly, depth data captured by sensors like LiDAR or Time-of-Flight cameras often appear sparse, limited by factors such as sensor range, occlusions, material reflectivity, and environmental conditions. At this juncture, the task of single-image depth completion emerges as a critical challenge. The objective is to reconstruct a comprehensive depth image from sparse depth data and RGB images, which presents numerous technical challenges.\\
%
Depth completion techniques that utilize a single sparse depth and an RGB image to estimate a dense depth image have recently shown significant advancements. These methods~\cite{CSPN, PENet, DySPN, NLSPN, DeepLiDAR, GuideNet, RigNet, CompletionFormer, TWISE, SpAgNet,guideformer}, rooted in supervised learning paradigms, typically require ground-truth annotations to function effectively. Although potent in controlled environments, these supervised approaches encounter challenges when faced with real-world scenarios' natural variability and unpredictability, where accurate ground-truth data are often scarce or unavailable.
Simply using parameters learned from other training datasets will cause severe performance degradation due to the distribution discrepancy between the two datasets.\\
To overcome the challenges posed by the domain gap and the lack of ground-truth data, several unsupervised depth completion methods have been developed~\cite{MaSelf, DFuseNet, DesNet}. These methods typically optimize the depth estimation by minimizing reprojection errors derived from multi-view images. Although these approaches address some of the limitations that supervised methods face, they heavily depend on auxiliary inputs such as stereo images or monocular video sequences. \\
\begin{figure*}[t]
  \centering
  \newcommand{\colw}{0.19\textwidth}
  \newcommand{\twotile}[4]{%
    \begin{minipage}[t]{\colw}\centering
      \includegraphics[width=\linewidth,page=#2]{#1}\par\vspace{2pt}
      \includegraphics[width=\linewidth,page=#3]{#1}\par\vspace{2pt}
      {\footnotesize #4}%
    \end{minipage}%
  }

  \twotile{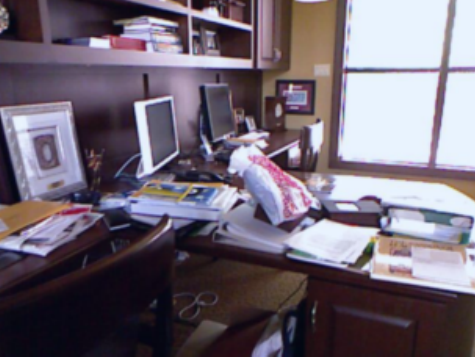}{1}{2}{(a) RGB}\hfill
  \twotile{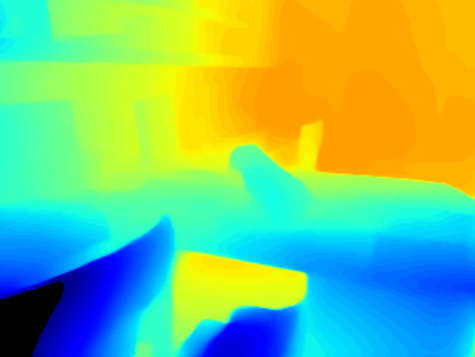}{1}{2}
        {\footnotesize \shortstack[c]{(b) Global-scale\\ affine transform}}\hfill
  \twotile{figure_assets/affine_figure1.pdf}{7}{8}
        {\footnotesize \shortstack[c]{(c) Segment-wise\\ affine transform}}\hfill
  \twotile{figure_assets/affine_figure1.pdf}{5}{6}{(d) \textbf{Ours}}\hfill
  \twotile{figure_assets/affine_figure1.pdf}{3}{4}{(e) Ground truth}

  \caption{Qualitative comparison of the depth map with the affine-transformed results of depth maps estimated by monocular depth estimators.}
  \label{fig:affine}
  \vspace{-2mm}
\end{figure*}
Meanwhile, a monocular depth estimation~(MDE) model seeks to infer relative depth from a single RGB image. Recent works in MDE~\cite{zoedepth, towards_zeroshot, MiDaS, DepthAnything, Metric3D} have significantly improved zero-shot depth estimation using a large number of parameters and large-scale datasets. These methodologies proficiently generate plausible relative depth maps from any given single RGB image, showcasing the potential of deep learning to interpret complex visual contexts.
However, as some works~\cite{miangoleh2021boosting, song2023darf, zhang2022hierarchical} have pointed out, MDE often fails to measure the depth difference between two objects accurately and faces challenges when determining whether the surface is concave, convex, or planar.


\begin{table*}[h]
    \centering

    \caption{Comparison of scale invariant metric (SILog).
    The values are multiplied by 100.}
    \begin{tabular}{lcc}
        \toprule
        Method & SILog$_\downarrow$ \\ 
        \midrule
        DepthAnything with global affine transformation & 184.87 \\ 
        DepthAnything with segment-wise affine transformation & 1.754 \\ 
        \midrule
        Ours & \textbf{0.022} \\ 
        \bottomrule
    \end{tabular}

    \label{tab:silog_segment}
\end{table*}

To this end, we introduce \textbf{StarryGazer}, a novel framework that effectively exploits and enhances the domain-agnostic capabilities of MDE models. Operating independently of ground-truth depth data, this framework is specifically designed to adapt to the diverse conditions typical of real-world environments, providing a robust solution for depth completion tasks without the constraints of dense depth data.
From the observation that the MDE models show prominent depth estimation accuracy within each object, we segment the depth map with a pre-trained large segmentation model and rescale the depth values of each segment independently to generate synthetic ground-truth. Then, the synthetic sparse depth map is generated by randomly sampling the points from the synthetic depth map. Then, the refinement network is trained to estimate the synthetic GT from the sparse depth map, MDE map, and RGB image. The parameters for rescaling each segment change for every iteration to widen the manifold of the synthetic depth maps and enable accurate metric depth estimation from the given sparse depth and image during inference.

To validate the superiority of our model, we apply an affine transformation to the MDE~\cite{DepthAnything} depth map to match the statistics of the MDE depth map to the given sparse depth map and evaluate the scale-invariant accuracy (SILog~\cite{SILog}) in Table~\ref{tab:silog_segment}.
The global transformed depth map, where all values are multiplied and added by a uniform scalar and bias, shows high SILog.
The SILog becomes much lower when the affine transforms are applied to each segment independently.
This validates the observation that more errors are produced from the relative positions of the objects than from inside each object.
However, as shown in Figure~\ref{fig:affine}, segment-wise transformed results include erroneous regions where the area of the segment is too small to get sufficient information from the sparse depth.
StarryGazer shows the best SILog and qualitative depth maps.
Extensive experiments validate its strong practical applicability, demonstrating competitive performance on real-world datasets and surpassing large-scale pretrained monocular depth estimation models on out-of-domain datasets~\cite{ClearGrasp, RHD}. Our main contributions can be summarized as follows: 
%
\begin{itemize}
\item We introduce StarryGazer, a domain-agnostic single-depth image completion method that does not require ground-truth depth data by exploiting the power of large monocular depth estimation models.
\item We devise a novel technique for generating synthetic ground-truth from segmented and rescaled relative depth maps, facilitating the training of our refinement network using only RGB and sparse depth inputs.
\item Our method achieves overwhelming performance in various datasets compared to existing unsupervised depth completion methods and a simple transformation of monocular depth estimation models.

\end{itemize}

\section{Related Works}
\label{Related Works}
\paragraph{Deep monocular depth estimation.}
Monocular depth estimation (MDE) has transitioned from its inception using ground-truth-dependent supervised methods, such as convolutional neural networks~(CNNs) with scale-invariant loss \cite{Eigen}, to sophisticated architectures incorporating conditional random fields \cite{Liu, newcrfs}, regression forests \cite{IEBins,Regression_forest}, transformer-based encoders \cite{zoedepth, Swin-Depth}, and diffusion model~\cite{Marigold}. The shift toward self-supervision aims to reduce annotation reliance, leveraging strategies like photometric consistency \cite{garg} and mixed reconstruction losses \cite{monodepth}. Recent advances focus on zero-shot depth estimation capable of handling images in varied domains without specific training \cite{zoedepth, towards_zeroshot, metric3dv2, unidepth, Metric3D, depth_anything_v2}, although these methods often struggle with generalization when predicting absolute depths. Our approach combines a domain-agnostic relative depth estimator with a robust absolute depth model to improve accuracy and generalizability in depth image estimation.

\paragraph{Supervised learning for depth completion.}
\label{Superivsed_Depth}
Depth image completion, the process of estimating dense depth images from sparse data, is crucial for applications such as autonomous navigation. 
The introduction of image-guided techniques marked a significant advancement, using additional modalities such as RGB images~\cite{PENet, RigNet}, semantic images~\cite{SemAttNet, CycleCons}, and surface normals~\cite{DeepLiDAR} to enrich and refine depth estimation.
Recent innovations have evolved to include multi-branch networks~\cite{DenseLiDAR, CrossGuidance, MSG-CHN, Depth-Normal, DenseDP, AdaContext} and spatial propagation networks~\cite{CSPN, CSPN++, DySPN, NLSPN, park2024depth}, which integrate these multi-modal inputs through advanced fusion techniques, enhancing the accuracy and structural integrity of the resulting depth images.
%
Despite rapid evolution in the field, these supervised methods still suffer from performance degradation when applied outside their training domain, reflecting an ongoing challenge of domain-agnostic depth completion.

\paragraph{Unsupervised learning for depth completion.}
\label{Unsupervised_Depth}
Unsupervised depth completion methods have emerged as a response to the limitations observed in supervised methods, particularly their performance degradation when applied outside their training domain. These methods~\cite{WongDistillation, Project2Adapt, MaSelf, DFuseNet, KBNet, VOICED, ScaffNet, WongAdaptive, Depth-Normal, DesNet, DenseDP} eliminate the need for ground-truth supervision during training. Instead, they used stereo images or monocular videos to compute photometric reprojection errors, minimizing discrepancies between input images and their reconstructions from other perspectives~\cite{DFuseNet, DenseDP, MaSelf, VOICED, ScaffNet, WongAdaptive}.
In particular, Wong \textit{et al.}~\cite{WongAdaptive} address occlusions by applying adaptive optimization to reduce penalties in these regions, while KBNet~\cite{KBNet} introduces a calibrated back projection network that enhances depth estimation accuracy by projecting image pixels onto 3D scenes.
Despite their efficacy in handling real-world data, these methods still rely on auxiliary inputs, such as stereo images or monocular video sequences, to function effectively.
In contrast, our approach aims to advance depth completion without ground-truth data or additional visual input, thereby overcoming the limitations of current unsupervised methods.

\begin{figure*}[ht]
   \centering
   \includegraphics[width=1\linewidth,page=5]{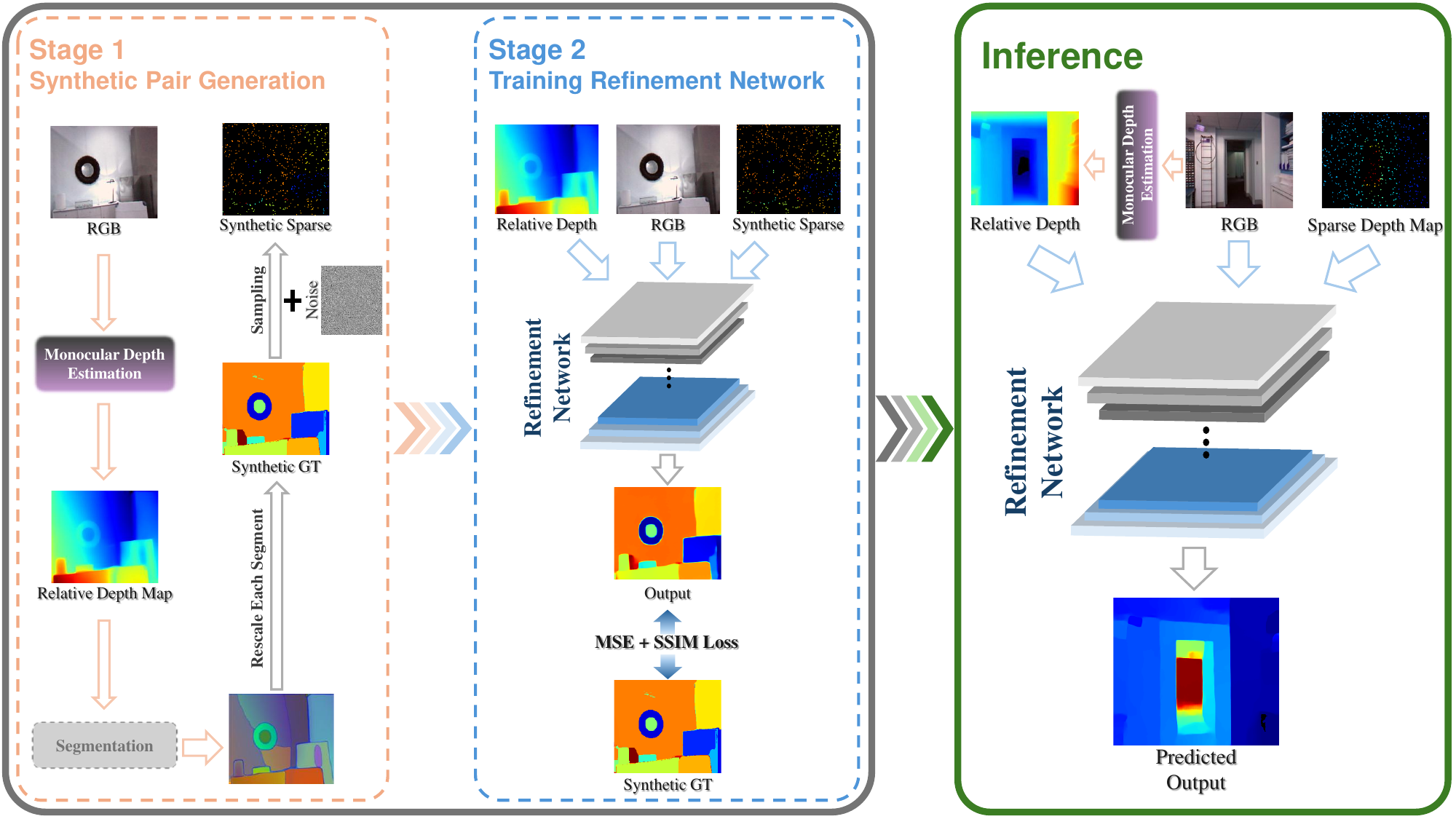}
   \caption{\textbf{Overall Pipeline of StarryGazer.} The training pipeline involves two stages: stage 1 generates synthetic pairs, while stage 2 trains the refinement network using the synthetic pairs. The trained refinement network uses real RGB images and sparse depth inputs to estimate dense depth maps during inference.}
   \label{fig:pipeline}
   \vspace{-4mm}
\end{figure*}

\section{Method}
\label{Method}
Our method, StarryGazer, leverages the large generalizable capacity from a pre-trained monocular depth estimation network to estimate depth from RGB images, circumventing the need for ground-truth depth images during training. As illustrated in Figure~\ref{fig:pipeline}, we first estimate a relative depth map from an off-the-shelf deep monocular depth estimation network and then segment the map into several classes. Then, a synthetic ground-truth depth is generated by adaptively rescaling the values of the depth map in a segment-wise manner. We sample the point from the synthetic ground truth to produce the corresponding synthetic sparse image. The refinement network is trained using the synthetic pairs. This framework introduces realistic depth variability, preparing the refinement model to perform robustly across diverse real-world scenes.
\subsection{Monocular Depth Estimation}
\label{Monocular Depth Estimation}
The relative depth images are generated using an existing deep monocular depth estimation network. The process is defined as follows:
\begin{equation}
    D_{\text{rel}} = f_{\text{MDE}}(I_{\text{RGB}}), 
\end{equation}
where \(D_{\text{rel}}\) represents the relative depth image produced, \(I_{\text{RGB}}\) denotes the input RGB image and \(f_{\text{MDE}}\) is the monocular depth estimation model. 
The MDE model captures geometric and semantic cues from a single image by leveraging a large-scale dataset during training.

\subsection{Generating Synthetic Pairs}
\label{Generating Synthetic Datasets}
A crucial part of our framework is the generation of synthetic pairs to train the refinement network. 
Although the MDE model tends to estimate the relative depths in the single object accurately, many errors occur at the boundary between the objects where the depth values change rapidly~\cite{chawla2021error}.
Therefore, we segment objects in the depth map and then independently rescale the depth values of each segment to add diversity to the depth relationships between objects while maintaining the depth relationships within the objects.
Using the off-the-shelf segmentation model, each relative depth image is normalized and segmented to identify distinct regions: 
\begin{equation}
    M_{\text{seg}} = f_{\text{seg}}\left(\frac{D_{\text{rel}} - \min(D_{\text{rel}})}{\max(D_{\text{rel}}) - \min(D_{\text{rel}})}\right),
\end{equation}
where \( M_{\text{seg}} \) denotes the segmentation map for the corresponding image and \( f_{\text{seg}} \) is an off-the-shelf semantic segmentation model. The segmentation model discriminates regions based on the normalized depth values. After segmentation, each identified region undergoes a rescaling process. The rescaling is performed adaptively, taking into account the local depth statistics within each region to simulate real-world depth variations more accurately:
\begin{equation}    D_{\text{sg}} = \sum_{k} \left( \alpha_k \cdot (D_{\text{rel}} \odot M_{\text{seg}}[ k]) + \beta_k \right),
\end{equation}
where $D_{\text{sg}}$ refers to synthetic ground-truth depth and \( \alpha_k \) and \( \beta_k \) are rescaling factors of the relative depth values within the region \( k \). Here, $M_{\text{seg}}[ k]$ denotes the segmentation mask for the $k$-th segment, which is set to 1, while all other segments are set to 0. Typically, \( \alpha_k \) is chosen to match the scale of depth variations observed in sparse depth data, while \( \beta_k \) adjusts the baseline depth to align with the observed minima and maxima within the region.
The operation \( \odot \) denotes element-wise multiplication, ensuring that the rescaling is applied independently for every segmented region. The adaptive rescaling factors \( \alpha_k \) and \( \beta_k\) are determined as follows:
\begin{align}
    \alpha_k &= \frac{2 \cdot \text{mean}(D_{\text{s}})}
                     {\text{mean}(D_{\text{rel}} \odot M_{\text{seg}}[k])}, \\
    \beta_k &\sim U\!\left(\min(D_{\text{s}}), \max(D_{\text{s}})\right)
               - \text{mean}(D_{\text{s}}),
\label{eq:alpha_beta}
\end{align}
where \(U(a, b)\) denotes the uniform distribution over the interval \([a, b]\) and $D_{\text{s}}$ is original sparse depth image.
These scaling adjustments ensure that the synthetic ground-truth images not only retain the structural integrity of the scene but also increase depth variability to encompass real-world depth distribution.
After rescaling, gaps or holes within the synthetic images where depth data might be missing due to incomplete segmentation masks are filled using an average filter. This process enhances the continuity of the synthetic depth images, which is crucial for training the refinement model. 
Then, synthetic sparse images are produced by applying the original sparse image as a mask, adding Gaussian noise to complement sensor inaccuracies robustly.
\begin{equation}
D_{\text{ss}} = (D_{\text{sg}}  + \mathcal{N}(0, \sigma^2)) \odot M_\text{s},
\end{equation}
where $D_{\text{ss}}$, $D_{\text{sg}}$ refers to synthetic sparse depth image and synthetic ground-truth depth image, respectively, and $M_\text{s}$ refers to the mask obtained from the original sparse depth map for $I_{RGB}$.
Adding Gaussian noise preserves the sparsity pattern and introduces realistic variability, enhancing the model performance with real sparse-depth images.

\subsection{Training Refinement Network}
\label{Loss Functions}
We train the refinement network from the synthetic pairs $(D_{sg}, D_{ss})$.
The refinement network receives the RGB image ($I_{RGB}$), the synthetic sparse map ($D_{ss}$) and the relative depth map ($D_{rel}$) and is trained to estimate the corresponding synthetic ground truth ($D_{sg}$).
Our training objective combines mean squared error~(MSE) and structural similarity index measure~(SSIM) loss functions, aiming to maximize the pixel-wise accuracy of the predicted depth images relative to the synthetic ground-truth. The MSE loss function calculates the mean squared difference between predicted and ground-truth depth values, defined as:
\begin{equation}
\mathcal{L}^{\text{mse}}(D_{\text{pred}}, D_{\text{sg}}) = \frac{1}{N} \sum_{i=1}^{N} (D_{\text{pred}}[i] - D_{\text{sg}}[i])^2,
\end{equation}
where \(N\) is the number of pixels in the depth image, $D_{\text{pred}}$ is the depth image predicted by the model, and $D_{\text{sg}}$  represents the synthetic ground-truth depth image.

Meanwhile, the SSIM loss function is used to improve the perceptual quality of predicted depth images by focusing on structural information, which is crucial to human visual perception. However, several works~\cite{mustafa2022training,zhao2016loss} have shown that SSIM loss is also helpful to improve the accuracy of results as well as perceptual quality. This SSIM loss is defined as:
\begin{equation}
\mathcal{L}^{\text{ssim}}(D_{\text{pred}}, D_{\text{sg}}) = 1 - \frac{1}{M} \sum_{j=1}^{M} ({SSIM\,map})_j,
\end{equation}
where \(M\) is the total number of local patches within the image, and ${SSIM\,map}$ is computed as:
\begin{equation}
    SSIM \, map = \frac{(2 \cdot \mu_{\text{pred}} \cdot  \mu_{\text{sg}} + C_1)(2 \cdot \sigma_{\text{pred,sg}} + C_2)}{(\mu_{\text{pred}}^2 + \mu_{\text{sg}}^2 + C_1)(\sigma_{\text{pred}}^2 + \sigma_{\text{sg}}^2 + C_2)},
\end{equation}
where \(\mu_{\text{pred}}\) and \(\mu_{\text{sg}}\) are the local means, and \(\sigma_{\text{pred}}^2\), \(\sigma_{\text{sg}}^2\), and \(\sigma_{\text{pred,sg}}\) are the local variances and covariance, computed using an averaging filter with a kernel size of 11.
\(C_1\) and \(C_2\) are constants that stabilize the division with weak denominator, defined as \(C_1 = (k_1 L)^2\) and \(C_2 = (k_2 L)^2\), with \(L\) being the dynamic range of the pixel-values (typically set to 1.0), and \(k_1 = 0.01\), \(k_2 = 0.03\).

The overall loss function used to train our model is a weighted sum of the MSE and SSIM losses:
\begin{equation}
    \mathcal{L}^{\text{total}} = \mathcal{L}^{\text{mse}}  + \lambda \cdot \mathcal{L}^{\text{ssim}},
\end{equation}
where \(\lambda\) is a hyperparameter. This combination of two loss terms yields higher depth estimation accuracy, which will be analyzed in detail in Section~\ref{sec:ablation}.

\section{Experiment}
\label{Experiment}

\newcommand{\cfoot}{\fontsize{7}{8}\selectfont}
\newcommand{\ctiny}{\fontsize{5.5}{7}\selectfont}
\subsection{Datasets}
\label{Datasets}
In this work, we utilize several well-established datasets to evaluate the performance and generalization of our depth completion model.\\
\textbf{NYU Depth V2~\cite{NYU}.} 
Comprising RGB and depth images extracted from video sequences of 464 indoor scenes, NYU Depth V2 has been a standard in depth completion benchmark. We follow conventional pre-processing from previous work~\cite{NLSPN, CompletionFormer} by resizing images from 640$\times$480 to 304$\times$228 and applying random sampling to create sparse depth images. We train on 50,000 training images and evaluate on 654 images from the official test set.\\
\textbf{KITTI Depth Completion (DC)~\cite{KITTIDC}.} Comprising 86,898 training samples along with 1,000 validation and 1,000 test samples, the KITTI Depth Completion dataset is a widely utilized real-world dataset for depth completion. The ground-truth is obtained from 11 LiDAR scans. Consistent with previous studies~\cite{ CompletionFormer}, we modify the image resolution to 1216$\times$240 by cropping the bottom center.\\
\textbf{ClearGrasp~\cite{ClearGrasp}.} ClearGrasp is a dataset that offers both synthetic and real-world scenarios focused on transparent objects, a challenging domain for depth sensing technologies.
Each segment of the ClearGrasp dataset, which comprises synthetic and real subsets, is further divided into training and novel sets. These novel sets contain classes that have not been encountered during the training phase, making the typical depth estimation model generate erroneous depth maps due to the large domain gap. \\

\subsection{Implementation Details}
\label{Implementation Details}
We use the DepthAnything VIT-S model~\cite{DepthAnything} and the Segment-Anything VIT-H model~\cite{SAM} for MDE and segmentation, respectively. We note that the datasets used for training DepthAnything and the datasets used for evaluation in this paper are entirely distinct, ensuring that no privileged information is involved in the pre-trained weights. For the refinement network, we slightly change the first CNN layer of NLSPN~\cite{NLSPN}, allowing the model to simultaneously receive RGB, sparse depth, and relative depth inputs. We use Adam optimizer~\cite{optim_adam} with a starting learning rate of 0.0001, adjusted by a step scheduler that reduces the rate by 0.8 for every five epochs. We set the loss weighting coefficient \(\lambda\) to 12 for training on the KITTI DC and 3 for the other datasets. We use MAE, RMSE, iMAE, and iRMSE for evaluation metrics. Experiments are conducted on Quadro RTX 8000 GPUs using PyTorch 1.10.1 \cite{paszke2019pytorch}. More details are provided in the supplementary material.

\subsection{Evaluation on Various Datasets}

\paragraph{Baseline: applying affine transformations to MDE depth maps.}
Since StarryGazer utilizes a large MDE model to train the refinement network, we include the results of the MDE model as baselines for evaluation.
However, the MDE model does not utilize the sparse absolute depth map information, and naively comparing the MDE depth maps with our results cannot be considered fair.
Thus, we apply a simple mathematical transformation, an affine transformation, to the MDE depth map to match the statistics of the sparse depth map values.
Specifically, the affine transformation is applied in two types: global scale and segment-wise.
For the global affine transformation case, a single scaling factor and bias are multiplied and added throughout the entire depth map.
The two values are determined to minimize the mean squared distance from absolute sparse depth values using the pseudo-inverse matrix.
The segment-wise affine transformation applies the affine transformation independently to each segment estimated by the segmentation model ($f_{seg}$).
For the segment that does not include any sparse depth points, we just use the MDE output value since there is no information about the statistics of the segment.

\begin{figure*}[t]
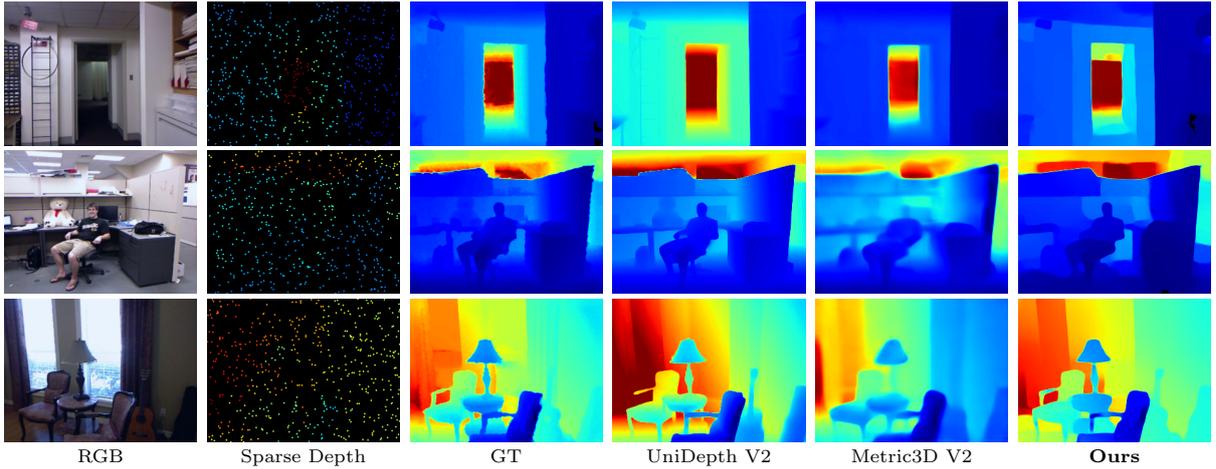

  \centering
  \setlength{\tabcolsep}{2pt}   
  \renewcommand{\arraystretch}{0} 
  \newcommand{\NYUtwo}[1]{\includegraphics[width=0.158\textwidth,page=#1]{figure_assets/NYU_qual2.pdf}}
  \newcommand{\NYUthree}[1]{\includegraphics[width=0.158\textwidth,page=#1]{figure_assets/NYU_qual3.pdf}}

  \begin{tabular}{@{}*{6}{c}@{}}
    \NYUtwo{10} & \NYUtwo{16} & \NYUtwo{1} & \NYUthree{3} & \NYUthree{5} & \NYUtwo{19} \\
    [2pt]
    \NYUtwo{11} & \NYUtwo{17} & \NYUtwo{2} & \NYUthree{2} & \NYUthree{4} & \NYUtwo{20} \\
    [2pt]
    \NYUtwo{12} & \NYUtwo{18} & \NYUtwo{3} & \NYUthree{1} & \NYUthree{6} & \NYUtwo{21} \\
    [3pt]
    \footnotesize RGB &
    \footnotesize Sparse Depth &
    \footnotesize GT &
    \footnotesize UniDepth V2 &
    \footnotesize Metric3D V2 &
    \footnotesize \textbf{Ours}
  \end{tabular}

  \caption{Qualitative comparison on NYU Depth V2 dataset.}
  \label{fig:nyu_qual}
  \vspace{-2mm}
\end{figure*}

\begin{table*}[t]
\centering
\caption{Quantitative results on the NYU Depth V2.}
\label{tab:nyu}

\small
\setlength{\tabcolsep}{3pt}  

\begin{tabular}{c|c|cc|cc}
    \toprule
    \multicolumn{2}{c|}{Method} & MAE (m)$\downarrow$ & RMSE (m)$\downarrow$ & iMAE (1/km)$\downarrow$ & iRMSE (1/km)$\downarrow$ \\
    \midrule
    \parbox[t]{2mm}{\multirow{5}{*}{\rotatebox[origin=c]{90}{Unsupervised}}}
     & SynthProj~\cite{Project2Adapt}            & 0.134 & 0.235 & 29.84  & 57.13 \\
     & VOICED~\cite{VOICED}               & 0.128 & 0.228 & 28.89  & 54.70 \\
     & ScaffNet~\cite{ScaffNet}             & 0.117 & 0.199 & 24.89  & 44.06 \\
     & KBNet~\cite{KBNet}                & 0.106 & 0.198 & 21.37  & 42.74 \\
     & DesNet~\cite{DesNet}               & 0.103 & 0.188 & 21.44  & 38.57 \\
    \midrule
    \parbox[t]{2mm}{\multirow{6}{*}{\rotatebox[origin=c]{90}{MDE}}}
     & DA~\cite{DepthAnything} (Global)  & 0.334 & 0.434 & 193.268 & 5549.994 \\
     & DA~\cite{DepthAnything} (Segment) & 0.226 & 0.514 &  38.445 &  164.335 \\
     & Metric3D~V2~\cite{metric3dv2} (Global)       & 1.004 & 1.037 & 337.330 &  368.462 \\
     & Metric3D~V2~\cite{metric3dv2} (Segment)      & 1.031 & 1.071 & 345.631 &  381.167 \\
     & UniDepth~V2~\cite{unidepth} (Global)    & 0.828 & 0.941 & 115.869 &  138.387 \\
     & UniDepth~V2~\cite{unidepth} (Segment)   & 0.952 & 1.100 & 142.706 &  180.426 \\
    \midrule
    \multicolumn{2}{c|}{Ours} & \textbf{0.100} & \textbf{0.171} & \textbf{18.78} & \textbf{35.93} \\
    \bottomrule
\end{tabular}
\end{table*}

\subsubsection{NYU V2 Dataset}
This section presents both quantitative and qualitative evaluations of our method on the NYU Depth V2 dataset, a widely recognized indoor dataset for depth completion. To ensure a fair comparison, we assume that no ground-truth depth is available during training, focusing exclusively on unsupervised methods.
Table~\ref{tab:nyu} outlines the performance of StarryGazer compared to existing approaches. 
The unsupervised methods show inferior quality to our results, highlighting the effectiveness of exploiting the prior knowledge from MDE models.
For affine transformed MDE maps, both global and segment-wise transformations show poor results.
Global transformation cannot correct the inaccuracy of the depth difference between objects in MDE, and cannot reduce the error effectively.
In the case of segment-wise transformation, since the sparse depth value of each segment is small, fitting with only the depth information within each segment can form an incorrect distribution with huge errors.
Our refinement network, in contrast, shows the best results by appropriately manipulating the depth map from MDE using the information from the global sparse map and the RGB image.
Figure~\ref{fig:nyu_qual} visually compares our method with other unsupervised techniques. Although the MDE model effectively captures semantic details and object shapes, it fails to estimate the relative position among the entities, often predicting near- and far-objects in opposite directions. 
Our method shows considerable qualitative improvements over these models and produces smoother depth images, approaching the quality of actual ground-truth. These visual assessments highlight the efficacy of our approach in generating realistic depth fields from sparse data, underlining its potential for robust real-world applications.

\begin{table*}[t]
\centering
\caption{Quantitative results on the KITTI DC dataset.}
\label{tab:kittiDC}

\small
\setlength{\tabcolsep}{2pt} 

\begin{tabular}{c|c|cc|cc}
\toprule
\multicolumn{2}{c|}{Method} & MAE (mm)$\downarrow$ & RMSE (mm)$\downarrow$ & iMAE (1/km)$\downarrow$ & iRMSE (1/km)$\downarrow$ \\
\midrule
\parbox[t]{2mm}{\multirow{8}{*}{\rotatebox[origin=c]{90}{Unsupervised}}}
 & SS-S2D~\cite{MaSelf}        & 350.32 & 1299.85 & 1.57 & 4.07 \\
 & DFuseNet~\cite{DFuseNet}      & 429.93 & 1206.66 & 1.79 & 3.62 \\
 & DDP~\cite{DenseDP}           & 343.46 & 1263.19 & 1.32 & 3.58 \\ 
 & VOICED~\cite{VOICED}        & 299.41 & 1169.97 & 1.20 & 3.56 \\
 & AdaFrame~\cite{WongAdaptive}      & 291.62 & 1125.67 & 1.16 & 3.32 \\
 & SynthProj~\cite{Project2Adapt}     & 280.42 & 1095.26 & 1.19 & 3.53 \\
 & ScaffNet~\cite{ScaffNet}      & 280.76 & 1121.93 & 1.15 & 3.30 \\
 & KBNet~\cite{KBNet}         & 258.36 & 1068.07 & \textbf{1.03} & 3.01 \\
\midrule
\parbox[t]{2mm}{\multirow{6}{*}{\rotatebox[origin=c]{90}{MDE}}}
 & DA~\cite{DepthAnything} (global)  & 3877.73 & 5520.75 & 147.69 & 6906.60 \\
 & DA~\cite{DepthAnything} (segment) & 1794.08 & 3332.65 &  15.60 &   35.54 \\
 & Metric3D V2~\cite{metric3dv2} (global)    & 3856.70 & 7903.73 &  51.67 &   76.48 \\
 & Metric3D V2~\cite{metric3dv2} (segment)   & 3257.82 & 4714.90 &  38.91 &   47.48 \\
 & UniDepth V2~\cite{unidepth} (global)    & 6155.49 & 7783.78 &  31.92 &   35.87 \\
 & UniDepth V2~\cite{unidepth} (segment)   & 7060.72 & 9595.91 &  36.16 &   44.95 \\
\midrule
\multicolumn{2}{c|}{Ours} & \textbf{242.44} & \textbf{1061.43} & 1.67 & \textbf{2.81} \\
\bottomrule
\end{tabular}

\end{table*}

\begin{figure*}[t]
\begin{center}
\renewcommand{\arraystretch}{0.2}
\begin{tabular}{@{}c@{\hskip 0.01\linewidth}c@{\hskip 0.01\linewidth}c@{\hskip 0.01\linewidth}c@{}}

\adjustbox{valign=c}{\small (a)} &
\includegraphics[width=0.3\linewidth, page=13]{figure_assets/KITTI_qual_new.pdf} &
\includegraphics[width=0.3\linewidth, page=14]{figure_assets/KITTI_qual_new.pdf} &
\includegraphics[width=0.3\linewidth, page=15]{figure_assets/KITTI_qual_new.pdf} \\[0.2cm]

\adjustbox{valign=c}{\small (b)} &
\includegraphics[width=0.3\linewidth, page=19]{figure_assets/KITTI_qual_new.pdf} &
\includegraphics[width=0.3\linewidth, page=20]{figure_assets/KITTI_qual_new.pdf} &
\includegraphics[width=0.3\linewidth, page=21]{figure_assets/KITTI_qual_new.pdf} \\[0.2cm]

\adjustbox{valign=c}{\small (c)} &
\includegraphics[width=0.3\linewidth, page=1]{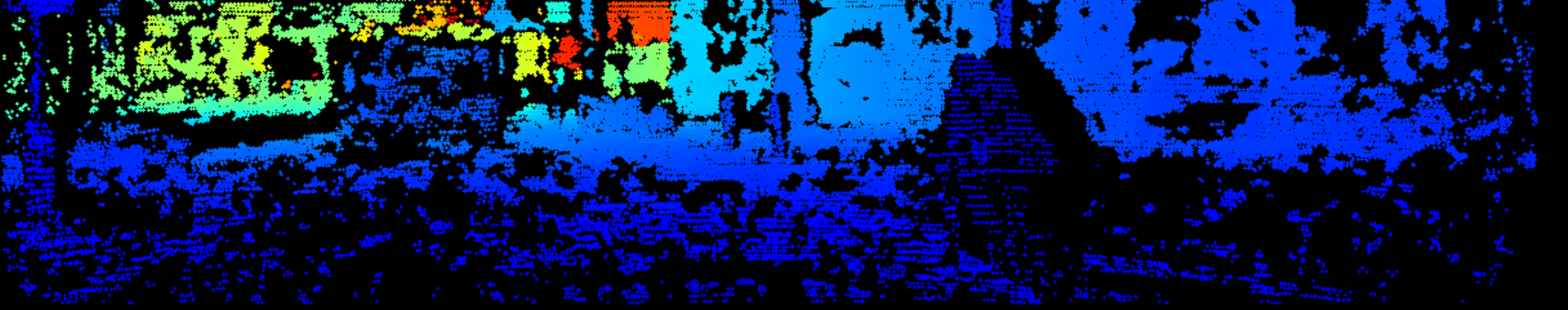} &
\includegraphics[width=0.3\linewidth, page=2]{figure_assets/KITTI_qual_new.pdf} &
\includegraphics[width=0.3\linewidth, page=3]{figure_assets/KITTI_qual_new.pdf} \\[0.2cm]

\adjustbox{valign=c}{\small (d)} &
\includegraphics[width=0.3\linewidth, page=16]{figure_assets/KITTI_qual_new.pdf} &
\includegraphics[width=0.3\linewidth, page=17]{figure_assets/KITTI_qual_new.pdf} &
\includegraphics[width=0.3\linewidth, page=18]{figure_assets/KITTI_qual_new.pdf} \\[0.2cm]

\adjustbox{valign=c}{\small (e)} &
\includegraphics[width=0.3\linewidth, page=4]{figure_assets/KITTI_qual_new.pdf} &
\includegraphics[width=0.3\linewidth, page=5]{figure_assets/KITTI_qual_new.pdf} &
\includegraphics[width=0.3\linewidth, page=6]{figure_assets/KITTI_qual_new.pdf} \\[0.2cm]

\adjustbox{valign=c}{\small (f)} &
\includegraphics[width=0.3\linewidth, page=7]{figure_assets/KITTI_qual_new.pdf} &
\includegraphics[width=0.3\linewidth, page=8]{figure_assets/KITTI_qual_new.pdf} &
\includegraphics[width=0.3\linewidth, page=9]{figure_assets/KITTI_qual_new.pdf} \\

\end{tabular}
\caption{Qualitative comparison on KITTI DC dataset. (a) RGB, (b) Sparse Depth, (c) Ground Truth (d) ScaffNet~\cite{ScaffNet}, (e) KBNet~\cite{KBNet}, (f) Ours.}
\label{fig:kitti_qual}
\end{center}
\vspace{-5mm}
\end{figure*}

\subsubsection{KITTI DC Dataset}
We demonstrate the practical applicability of StarryGazer in real-world outdoor scenarios, particularly under conditions where sparse depth data is unevenly distributed. Unlike other unsupervised methods that rely on auxiliary inputs such as stereo images or camera intrinsic matrices, as mentioned in Section~\ref{Related Works}, our approach operates without such additional data.\\
Table~\ref{tab:kittiDC} compares our method quantitatively with other unsupervised learning approaches on the KITTI DC dataset. Our method demonstrates superior performance in terms of MAE and RMSE, achieving the lowest scores of 242.44 mm and 1061.43 mm, respectively. This indicates an improvement over competing methods, confirming the effectiveness of our approach in handling diverse driving scenarios. However, in terms of iMAE, our method exhibits a slight underperformance, registering 1.67 compared to the best score of 1.03 by KBNet~\cite{KBNet}. This happens because our refinement network is not specially designed for sparsity robustness and produces higher errors at closer ranges, where the errors are more penalized. Meanwhile, our method still shows superiority on iRMSE, which implies that errors in these regions are consistently distributed without significant outliers.
Despite this minor limitation, the overall results strongly validate the robustness and reliability of our method in real-world applications.
Also, similar to the result in NYU V2, our model outperforms transformed MDE results with a large margin by exploiting the information from the input image.
Figure~\ref{fig:kitti_qual} presents the qualitative comparison on the KITTI DC dataset. The visual results clearly demonstrate that our method not only estimates smoother depth images but also recovers local details more effectively than competing unsupervised methods. These results demonstrate our model's ability to handle complex outdoor scenes, preserving both major structures and finer details essential for real-world applications.

\subsubsection{ClearGrasp Dataset}
\begin{table}[t]
\centering
\caption{Quantitative comparison on ClearGrasp dataset.}
\label{tab:ood}

\small
\setlength{\tabcolsep}{0.2pt}

\begin{tabular}{c|p{4cm}|cc}
\toprule
\multicolumn{2}{c|}{Method} & MAE (m)$\downarrow$ & RMSE (m)$\downarrow$ \\
\midrule
\parbox[t]{3mm}{\multirow{7}{*}{\rotatebox[origin=c]{90}{Supervised}}}
 & DenseDepth~\cite{DenseDepth}      & 0.260 & 0.270 \\
 & DeepCompletion~\cite{DeepCompletion}  & 0.045 & 0.054 \\
 & NLSPN~\cite{NLSPN}                 & 0.106 & 0.132 \\
 & ClearGrasp~\cite{ClearGrasp}       & 0.038 & 0.044 \\
 & LocalImplicit~\cite{LocalImplicit} & 0.034 & 0.041 \\
 & TranspareNet~\cite{TranspareNet}   & 0.027 & \textbf{0.032} \\
 & Depth Prompting~\cite{park2024depth} & 0.462 & 0.530 \\
\midrule
\parbox[t]{3mm}{\multirow{6}{*}{\rotatebox[origin=c]{90}{MDE}}}
 & DA~\cite{DepthAnything}(Global)  & 0.077 & 0.120 \\
 & DA~\cite{DepthAnything}(Segment) & 0.065 & 0.139 \\
 & Metric3D~V2~\cite{metric3dv2}(Global)       & 0.230 & 0.261 \\
 & Metric3D~V2~\cite{metric3dv2}(Segment)      & 0.221 & 0.245 \\
 & UniDepth~V2~\cite{unidepth}(Global)         & 0.296 & 0.311 \\
 & UniDepth~V2~\cite{unidepth}(Segment)        & 0.301 & 0.332 \\
\midrule
\multicolumn{2}{c|}{Ours} & \textbf{0.025} & 0.062 \\
\bottomrule
\end{tabular}

\end{table}

To demonstrate the generalizability of our model in various domains, we performed evaluations on the ClearGrasp dataset. It has different image manifolds from typical datasets used for training supervised depth completion networks.
Obtaining ground-truth data is often challenging in real-world scenarios, and maintaining performance in unseen environments is particularly challenging. 
Our method's advantage is that it can be trained using synthetically generated pairs without relying on ground-truth data, making it particularly suitable for handling such challenges.
Table~\ref{tab:ood} provides a quantitative comparison of our method against supervised depth completion methods and transformed MDE methods.
Our method achieves the lowest errors on both datasets, underscoring its robustness in out-of-domain scenarios.\\ 
\begin{figure*}[t]
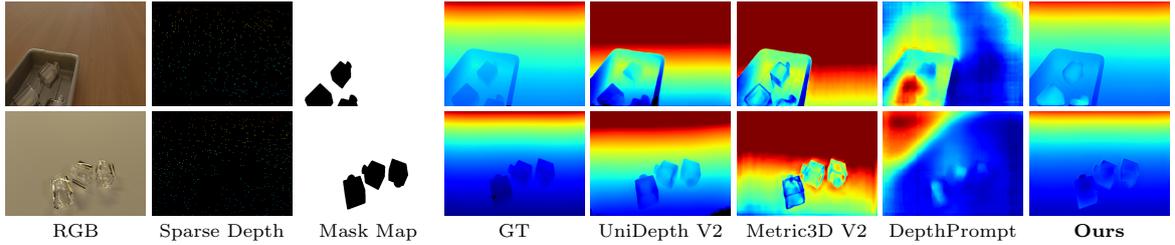

  \centering
  \setlength{\tabcolsep}{1.2pt}
  \renewcommand{\arraystretch}{0}

  \newcommand{\CG}[1]{%
    \includegraphics[width=0.115\textwidth,page=#1]{figure_assets/ClearGrasp_qual2.pdf}%
  }

  \begin{tabular}{@{}*{8}{c}@{}}
    \CG{11} & \CG{13} & \CG{5} & \CG{3} & \CG{15} & \CG{7} & \CG{1} & \CG{9} \\
    [2pt]
    \CG{12} & \CG{14} & \CG{6} & \CG{4} & \CG{16} & \CG{8} & \CG{2} & \CG{10} \\
    [3pt]
    \footnotesize RGB &
    \footnotesize Sparse Depth &
    \footnotesize Mask Map &
    \footnotesize GT &
    \footnotesize UniDepth V2 &
    \footnotesize Metric3D V2 &
    \footnotesize DepthPrompt &
    \footnotesize \textbf{Ours}
  \end{tabular}

  \caption{Qualitative comparison on ClearGrasp dataset. In this setting, the mask map restricts sparse depth data to non-masked areas, leaving no depth values for transparent objects.}
  \label{fig:ClearGrasp}
  \vspace{-2mm}
\end{figure*}

As shown in Figure~\ref{fig:ClearGrasp}, the mask map restricts sparse depth values to non-masked areas, leaving no depth points available for transparent objects. This setup poses a challenge for competing methods that struggle to estimate depth accurately, particularly in regions with transparent objects. In contrast, our method successfully estimates the depth values for these objects, producing results that are closely aligned with the ground-truth.

\newcommand{\chkmark}{\textcolor{blue}{\ding{51}}}
\newcommand{\crossmark}{\textcolor{red}{\ding{55}}}

\begin{table}[t]
\centering
\caption{Ablations on different values of $\alpha_k$ and $\beta_k$.}
\label{tab:ablation_alpha_beta}

\small
\setlength{\tabcolsep}{6pt} 

\begin{tabular}{cc|cc}
\toprule
$\alpha_k$ & $\beta_k$ & MAE (m)$\downarrow$ & RMSE (m)$\downarrow$ \\
\midrule
Eq.~\ref{eq:rand_alpha} & 0                         & 2.72  & 2.91 \\
Eq.~\ref{eq:rand_alpha} & Eq.~\ref{eq:alpha_beta}   & 0.234 & 0.326 \\
\midrule
Eq.~\ref{eq:alpha_beta} & Eq.~\ref{eq:alpha_beta}   & \textbf{0.100} & \textbf{0.171} \\
\bottomrule
\end{tabular}

\end{table}

\begin{table}[t]
\centering
\caption{Effect of $\mathcal{L}^{ssim}$ and noise addition.}
\label{tab:ablation_loss}

\small
\setlength{\tabcolsep}{6pt} 

\begin{tabular}{ccc|cc}
\toprule
$\mathcal{L}^{mse}$ & $\mathcal{L}^{ssim}$ & Noise & MAE (m)$\downarrow$ & RMSE (m)$\downarrow$ \\
\midrule
\chkmark & \crossmark & \crossmark & 0.132 & 0.212 \\
\chkmark & \chkmark   & \crossmark & 0.118 & 0.190 \\
\chkmark & \crossmark & \chkmark   & 0.112 & 0.192 \\
\midrule
\chkmark & \chkmark   & \chkmark   & \textbf{0.100} & \textbf{0.171} \\
\bottomrule
\end{tabular}

\end{table}
\subsection{Ablation Studies}
\label{sec:ablation}

\paragraph{Effect of $\alpha_k, \beta_k$ configuration for training.}
Since the refinement network is trained with synthetic pairs, the synthetic ground-truth should be similar to the original depth distribution.
We change the configuration of $\alpha$ and $\beta$ when generating synthetic ground-truth for training.
When we choose $\alpha$ in a stochastic manner, $\alpha$ is sampled from the following distribution:
\begin{equation}
    \alpha_k \sim U(\min(D_s), \max(D_s)).
\label{eq:rand_alpha}
\end{equation}
Table~\ref{tab:ablation_alpha_beta} shows the results by changing the distributions of $\alpha_k$ and $\beta_k$.
First, giving the fixed value $\beta_k$ severely degrades the quality.
The distribution discrepancy between synthetic GT and real GT remains when we use a fixed $\beta_k$ value.
On the contrary, as $\beta_k$ varies at each iteration, the distribution of synthetic GTs is expanded to include real GTs.
In case of $\alpha_k$, making $\alpha_k$ flexible will over-broaden the distribution of synthetic data, leading the model to be underfitted to such a large distribution.
Our configuration shows the best bias-variance tradeoff for designating the distribution of synthetic data.

\paragraph{Effect of $L^{ssim}$ and noise addition.}
We evaluate the impact on model performance of excluding $\mathcal{L}^{ssim}$ and adding synthetic noise during training, as shown in Table~\ref{tab:ablation_loss}.
The omission of $\mathcal{L}^{ssim}$ from the loss function led to a slight degradation in both MAE and RMSE, highlighting its crucial role in enhancing the smoothness and visual coherence of the depth images.
The absence of noise in the training process also resulted in a modest deterioration in performance, suggesting that the incorporation of noise aids in mimicking the real-world inaccuracies found in sensor data.
Removing both $\mathcal{L}^{ssim}$ and noise led to the most significant performance drop, underlining their combined importance in achieving optimal results. These findings underscore the necessity of both $\mathcal{L}^{ssim}$ for depth image refinement and synthetic noise for realistic training conditions, affirming their synergistic effect in improving the model's performance in depth completion tasks.

\section{Conclusion}
\label{Conclusion}
We introduce StarryGazer, a domain-agnostic approach for depth completion using monocular depth estimation models. The MDE depth maps are segmented and rescaled to generate synthetic sparse and dense depth pairs. The refinement network trained on the synthetic dataset estimates the final dense depth map from the RGB image and the sparse depth map.
The training framework leverages the strong prior of depth maps generated by the MDE model while fixing the error of the MDE estimated maps by augmenting the synthetic ground truth using segment-wise rescaling.

\noindent \textbf{Limitations and future works.}
While StarryGazer has shown promising results, the framework will fail where the synthetic ground-truth map cannot imitate the distribution of real depth maps. Future works can involve adaptive segmentation models or an improved rescaling policy that considers the RGB image to generate a more proper synthetic data distribution.

\section*{Statements and Declarations}
\noindent\textbf{Competing interests} 
The authors declare that they have no known competing financial interests or personal relationships that could have appeared to influence the work reported in this paper. The authors declare the following financial interests/personal relationships which may be considered as potential competing interests.

\noindent\textbf{Authors contributions}
S.H. and S.L. designed the method and conducted the experiments. S.H. implemented the framework and performed ablation studies. S.L. conducted comparative evaluations. K.M.L. supervised the project and provided critical feedback on the methodology and manuscript.




\begingroup
\scriptsize
\setlength{\bibsep}{0pt}
\bibliography{sn-bibliography}
\endgroup

\end{document}


\title[Article Title]{Supplementary Information {\it for} StarryGazer: Leveraging Monocular Depth Estimation Models for Domain-Agnostic Single Depth Image Completion}

\renewcommand{\thetable}{S\arabic{table}}
\renewcommand{\thefigure}{S\arabic{figure}}
\renewcommand{\theequation}{S\arabic{equation}}
\renewcommand{\thesection}{S\arabic{section}}


\author[1]{\fnm{Sangmin} \sur{Hong}}\email{mchiash2@snu.ac.kr}
\equalcont{These authors contributed equally to this work.}
\author[2]{\fnm{Suyoung} \sur{Lee}}\email{esw0116@snu.ac.kr}
\equalcont{These authors contributed equally to this work.}

\author*[1,2]{\fnm{Kyoung Mu} \sur{Lee}}\email{kyoungmu@snu.ac.kr}

\affil[1]{IPAI, Seoul National University, Seoul 08826, South Korea}
\affil[2]{Department of Electrical and Computer Engineering, ASRI, Seoul National University, Seoul 08826, South Korea}

\maketitle

\section{Evaluation Metrics}
We describe the formula for each evaluation metric used to assess performance.
We employ standard metrics commonly used in depth estimation and completion tasks.
We note that all metrics except $\delta_{\tau}$ are better when they become lower.

\begin{itemize}
    \setlength\itemsep{2mm}
    \item RMSE: {\tiny $\sqrt{\frac{1}{\left\vert \mathcal{V} \right\vert}\sum_{v \in \mathcal{V}}\left\vert\ d_{v}^{gt} - d_{v}^{pred}\ \right\vert^{2}}$}
    \item MAE: {\tiny $\frac{1}{\left\vert \mathcal{V} \right\vert}\sum_{v \in \mathcal{V}}\left\vert\ d_{v}^{gt} - d_{v}^{pred}\ \right\vert$}
    \item iRMSE: {\tiny $\sqrt{\frac{1}{\left\vert \mathcal{V} \right\vert}\sum_{v \in \mathcal{V}}\left\vert\ 1/d_{v}^{gt} - 1/d_{v}^{pred}\ \right\vert^{2}}$}
    \item iMAE: {\tiny $\frac{1}{\left\vert \mathcal{V} \right\vert}\sum_{v \in \mathcal{V}}\left\vert\ 1/d_{v}^{gt} - 1/d_{v}^{pred}\ \right\vert$}
    \item SILog $=
    \begin{aligned}[t]
    &\displaystyle \frac{1}{|\mathcal V|}\sum_{v\in\mathcal V}
      \bigl(\log d_v^{\text{pred}}-\log d_v^{\text{gt}}\bigr)^2\\[-2pt]
    &\displaystyle \qquad
      -\left(\frac{1}{|\mathcal V|}\sum_{v\in\mathcal V}
      \bigl(\log d_v^{\text{pred}}-\log d_v^{\text{gt}}\bigr)\right)^2
    \end{aligned}$ 
    \item Rel: {\tiny $\frac{1}{\left\vert \mathcal{V} \right\vert}\sum_{v \in \mathcal{V}}\left\vert\ (d_{v}^{gt} - d_{v}^{pred}) / d_{v}^{gt}\ \right\vert$}
    \item $\delta_{\tau}$: Percentage of pixels satisfying
    \hspace*{0.01\linewidth} {\tiny $\texttt{max}\left(\frac{d_{v}^{gt}}{d_{v}^{pred}}, \frac{d_{v}^{pred}}{d_{v}^{gt}}\right) < \tau$}
\end{itemize}

\section{Scale Invariant Metric in Global and Segment Scale}
\begin{table*}[h]
\centering
\caption{Comparison of scale invariant metric (SILog) in two scales (global, segment). The values are multiplied by 100.}
\label{tab:supp_silog}

\begin{tabular}{l|cc}
\toprule
Method & Global & Segment \\
\midrule
DepthAnything with global affine transformation      & 184.87 & 8.51 \\
DepthAnything with segment-wise affine transformation & 1.754  & 1.18 \\
\midrule
Ours                                                  & \textbf{0.022} & \textbf{0.023} \\
\bottomrule
\end{tabular}

\end{table*}

We show the superiority of our method over the simple combination of an affine transformation and an MDE model in Table 1. To further justify synthetic pair generation with random rescaling independently for each segment, we measure SILog metrics in two scales, global and segment, and report the results in Table~\ref{tab:supp_silog}.
The result of a typical SILog calculation between the estimated depth map and the ground truth depth map is shown in the `global' column, which is identical to Table 1.
For the segment scale, we first segment the MDE depth map with the segmentation network ($f_{seg}$), compute the SILog for each segment, and average the calculated SILog metrics. Since we measure scale invariant accuracy within each segment, depth error between segments is not calculated. The segment scale SILog is written in the `segment' column.

The significant reduction in SILog when computed segment-wise reveals a key insight: while the global scale of monocular depth predictions is unreliable, local regions often exhibit internally consistent depth structures. This observation provides strong empirical motivation for our proposed method—refining relative depth predictions by segmenting and rescaling each region individually to synthesize training data. By leveraging this piecewise geometric consistency, \textbf{StarryGazer} circumvents the need for ground-truth depth while achieving high accuracy across domains.

\section{Comparison with Additional Baseline Methods}
\begin{table*}[h]
    \centering
    \begin{minipage}[t]{0.4\linewidth}
    \caption{Quantitative Comparison with Supervised Depth Completion Methods on NYU Depth V2.}
    \resizebox{\linewidth}{!}{
    \begin{tabular}{c|cc}
        \toprule
        Method & RMSE $\downarrow$ & Rel $\downarrow$  \\
        \midrule
        CSPN~[\textcolor{blue}{1}] & $0.117$ & $0.016$ \\
        DeepLiDAR~[\textcolor{blue}{5}] & $0.115$ & $0.022$  \\
        GuideNet~[\textcolor{blue}{6}] & $0.101$ & $0.015$ \\
        NLSPN~[\textcolor{blue}{4}]  & $0.092$ & $0.012$  \\
        ACMNet~[\textcolor{blue}{45}]  & $0.105$ & $0.015$  \\
        TWISE~[\textcolor{blue}{9}]  & $0.097$ & $0.013$  \\
        RigNet~[\textcolor{blue}{7}]  & $0.090$ & $0.012$  \\ 
        DySPN~[\textcolor{blue}{3}] & $0.090$ & $0.012$ \\ 
        SpAgNet~[\textcolor{blue}{10}]& $0.114$ & $0.015$\\  
        CompletionFormer~[\textcolor{blue}{8}] & $0.090$ & $0.012$  \\  
        \midrule
        Ours & $0.171$ & $0.039$\\
        \bottomrule
    \end{tabular}
    }
    \label{tab:Sup_NYU}
    \end{minipage}
    \hspace{0.06\linewidth}
    \begin{minipage}[t]{0.4\linewidth}
    \caption{Quantitative Comparison with Supervised Depth Completion Methods on KITTI DC.}
    \resizebox{\linewidth}{!}{
    \begin{tabular}{c|cc}
        \toprule
        Method & MAE $\downarrow$ & RMSE $\downarrow$ \\
        \midrule
        CSPN~[\textcolor{blue}{1}]  & $0.279$ & $1.019$ \\
        DeepLiDAR~[\textcolor{blue}{5}] &  $0.226$ & $0.758$ \\
        GuideNet~[\textcolor{blue}{6}] &  $0.218$ & $0.736$ \\
        NLSPN~[\textcolor{blue}{4}] & $0.199$ & $0.741$ \\
        PENet~[\textcolor{blue}{2}] & $0.210$ & $0.730$ \\
        ACMNet~[\textcolor{blue}{45}] &  $0.206$ & $0.744$ \\
        TWISE~[\textcolor{blue}{9}]  & $0.195$ & $0.840$ \\
        RigNet~[\textcolor{blue}{7}]& $0.203$ & $0.712$ \\
        GuideFormer~[\textcolor{blue}{11}] & $0.207$ & $0.721$ \\  
        DySPN~[\textcolor{blue}{3}] &  $0.192$ & $0.709$ \\ 
        CompletionFormer~[\textcolor{blue}{8}] & $0.203$ & $0.708$ \\  \
        SemAttNet~[\textcolor{blue}{38}] & $0.205$ & $0.709$ \\  
        \midrule
        Ours & $0.242$ & $1.061$ \\
        \bottomrule
    \end{tabular}
    }
    \label{tab:Sup_KITTI}
    \end{minipage}
    \label{tab:Sup}
\end{table*}
\subsection{Supervised Depth Completion Methods}
We mostly compared our method with unsupervised depth completion approaches in the main manuscript.
Here, we list the results of several state-of-the-art supervised depth completion methods in Table~\ref{tab:Sup_NYU} and \ref{tab:Sup_KITTI} for NYU Depth V2 and KITTI DC dataset, respectively.
Considering that the supervised methods require ground truth dense depth maps when training the network, our method achieves reasonable performance to the supervised methods without using the ground truth.
\begin{table*}[ht]
    \centering
    \begin{minipage}[t]{0.45\linewidth}
    \centering
    \caption{Quantitative comparison with MDE methods on NYU Depth V2 dataset.}
    \resizebox{\linewidth}{!}{%
        \begin{tabular}{c|c|c|c}
            \toprule
            Method & RMSE $\downarrow$  & $\delta_{\substack{\text{1.25}}} \uparrow$ & Rel $\downarrow$  \\
            \midrule
            ZoeDepth~[\textcolor{blue}{15}]  &   $0.277$  & $0.953$ & $0.077$ \\
            ZeroDepth~[\textcolor{blue}{16}] &  $0.269$  & $0.954$ & $0.074$ \\
            NeWCRFs~[\textcolor{blue}{28}]  & $0.334$ & $0.922$ & $0.095$\\
            IEBins~[\textcolor{blue}{29}]&  $0.314$  & $0.936$ & $0.087$ \\
            Metric3D~[\textcolor{blue}{19}] &  $0.187$  & $0.987$ & $0.045$ \\
            DepthAnything~[\textcolor{blue}{18}] &  $0.206$  & $0.984$ & $0.056$ \\
            DepthAnything V2~[\textcolor{blue}{37}] &  $0.206$  & $0.979$ & $0.044$ \\
            \midrule
            Ours   & $\textbf{0.171}$  & $\textbf{0.999}$ & $\textbf{0.039}$ \\
            \bottomrule
        \end{tabular}
    }
    \label{tab:MDE_NYU}
    \end{minipage}
    \hspace{0.06\linewidth}
    \begin{minipage}[t]{0.45\linewidth}
    \centering
    \caption{Quantitative comparison with MDE methods on KITTI DC dataset.}
    \resizebox{\linewidth}{!}{%
        \begin{tabular}{c|c|c|c}
            \toprule
            Method  & RMSE $\downarrow$ & $\delta_{\substack{\text{1.25}}} \uparrow$ & Rel $\downarrow$  \\
            \midrule
            ZoeDepth~[\textcolor{blue}{15}] & $2.281$  & $0.971$ & $0.053$ \\
            ZeroDepth~[\textcolor{blue}{16}]  & $2.087$  & $0.968$ & $0.057$ \\
            NeWCRFs~[\textcolor{blue}{28}] & $2.129$ & $0.974$ & $0.052$ \\
            IEBins~[\textcolor{blue}{29}] & $2.011$  & $0.978$ & $0.050$ \\
            Metric3D~[\textcolor{blue}{19}]  & $1.766$  & $0.989$ & $0.039$ \\
            DepthAnything~[\textcolor{blue}{18}]  & $1.896$  & $0.982$ & $0.046$ \\
            DepthAnything V2~[\textcolor{blue}{37}] & $1.861$ & $0.983$ & $0.045$ \\
            \midrule
            Ours  & $\textbf{1.061}$  & $\textbf{0.995}$ & $\textbf{0.039}$ \\
            \bottomrule
        \end{tabular}
    }
    \label{tab:MDE_KITTI}
    \end{minipage}
    \label{tab:MDE}
\end{table*}
\subsection{Monocular Depth Estimation Methods}
As shown in Tables~\ref{tab:MDE_NYU} and \ref{tab:MDE_KITTI}, we present comparisons with several state-of-the-art MDE methods on the NYU Depth V2 and KITTI DC datasets.
Unlike MDE methods, which rely solely on RGB input, our approach leverages both RGB and sparse absolute depth data. Our method consistently outperforms the MDE methods in both datasets, achieving the lowest RMSE and Rel values, along with high $\delta_{\substack{\text{1.25}}}$ scores, clearly demonstrating the effectiveness of integrating sparse depth information to improve estimation accuracy. 
On the NYU Depth V2 dataset, our method achieves an RMSE of $0.171$ and a Rel of $0.039$, and on the KITTI DC dataset, it achieves an RMSE of $1.061$ and a Rel of $0.039$, outperforming the MDE methods.
These results highlight the importance of combining RGB and sparse depth inputs to produce more reliable and accurate depth estimations.

\section{Experiments on Rendered Handpose Dataset}


\begin{table}[h]
\centering
\caption{Quantitative comparison on Rendered Handpose.}
\label{tab:handpose}

\begin{tabular}{p{3.7cm}|p{1.4cm}|p{1.4cm}}
\toprule
Method & MAE~(m)$\downarrow$ & RMSE~(m)$\downarrow$ \\
\midrule
NLSPN~[\textcolor{blue}{4}]                & 0.485 & 0.563 \\
CompletionFormer~[\textcolor{blue}{8}]     & 0.443 & 0.524 \\
Depth Prompting~[\textcolor{blue}{47}]     & 0.700 & 0.797 \\
\midrule
Metric3D V2~[\textcolor{blue}{35}] (global) & 0.479 & 0.564 \\
Metric3D V2~[\textcolor{blue}{35}] (segment)& 0.447 & 0.532 \\
UniDepth V2~[\textcolor{blue}{36}] (global) & 0.407 & 0.526 \\
UniDepth V2~[\textcolor{blue}{36}] (segment)& 0.398 & 0.478 \\
\midrule
Ours & \textbf{0.352} & \textbf{0.408} \\
\bottomrule
\end{tabular}

\end{table}

\begin{figure*}[t]
  \centering
  \setlength{\tabcolsep}{2pt}           
  \renewcommand{\arraystretch}{0}       

  \newcommand{\RHD}[1]{%
    \includegraphics[width=0.128\textwidth,page=#1]{figure_assets/RHD_qual4.pdf}%
  }

  \begin{tabular}{@{}*{7}{c}@{}}
    \RHD{9}  & \RHD{11} & \RHD{3}  & \RHD{13} & \RHD{5}  & \RHD{1}  & \RHD{7}  \\
    [2pt]
    \RHD{10} & \RHD{12} & \RHD{4}  & \RHD{14} & \RHD{6}  & \RHD{2}  & \RHD{8}  \\
    [3pt]
    \footnotesize RGB &
    \footnotesize Sparse Depth &
    \footnotesize GT &
    \footnotesize UniDepth V2 &
    \footnotesize Metric3D V2 &
    \footnotesize DepthPrompting &
    \footnotesize \textbf{Ours}
  \end{tabular}

  \caption{Qualitative results on the Rendered Handpose dataset (out-of-domain). Sparse depth points exist only on the human figure; the background has no depth.}
  \label{fig:RHD}
  \vspace{-2mm}
\end{figure*}
As originally designed for hand pose estimation, the Rendered Handpose dataset~[\textcolor{blue}{25}] consists of RGB images, corresponding depth maps, segmentation masks, and key point locations. Since there is no depth information on the background of humans, we train the model only on the human region of the segmentation mask. It consists of 41,258 and 2,728 training and testing pairs, respectively.\\
The quantitative results in Table~\ref{tab:handpose} show that our method estimates much more accurate depth values than the unsupervised or MDE-based methods. Figure~\ref{fig:RHD} presents the visual results on the Rendered Handpose dataset. In this dataset, sparse depth points are available only in the human region, with no depth information provided for the background. The absence of background depth data makes it particularly challenging to estimate background depth values accurately. Nevertheless, our method accurately estimates depth values for points on the human figure, achieving a close approximation to the ground truth. In contrast, the competing method struggles to capture the fine details of the human figure. These results underscore our approach's superior out-of-domain generalization, highlighting the model's robustness.

\section{Additional Ablation Studies}

\begin{table*}[h]
\centering
\caption{Sparsity analysis on NYU Depth V2: Quantitative comparison with KBNet using MAE and RMSE across different numbers of sparse input points (50, 200, 500, 2000).}
\label{tab:sparsity_comparison}

\begin{tabular}{c|cc|cc|cc|cc}
\toprule
\# of Points & \multicolumn{2}{c|}{50} & \multicolumn{2}{c|}{200} & \multicolumn{2}{c|}{500} & \multicolumn{2}{c}{2000} \\
\midrule
Method & MAE$\downarrow$ & RMSE$\downarrow$ & MAE$\downarrow$ & RMSE$\downarrow$ & MAE$\downarrow$ & RMSE$\downarrow$ & MAE$\downarrow$ & RMSE$\downarrow$ \\
\midrule
KBNet~[\textcolor{blue}{50}] & 0.300 & 0.450 & 0.180 & 0.280 & 0.106 & 0.198 & 0.102 & 0.170 \\
\midrule
Ours & \textbf{0.256} & \textbf{0.400} & \textbf{0.150} & \textbf{0.240} & \textbf{0.100} & \textbf{0.171} & \textbf{0.097} & \textbf{0.146} \\
\bottomrule
\end{tabular}

\end{table*}

\begin{figure}[!t]
  \centering
  \setlength{\tabcolsep}{1.5pt}   
  \renewcommand{\arraystretch}{0} 

  \newcommand{\SQ}[1]{%
    \includegraphics[width=.238\columnwidth,page=#1]{figure_assets/Sparsity_qual.pdf}%
  }

  \begin{tabular}{@{}cccc@{}}
    \SQ{4}  & \SQ{1}  & \SQ{5}  & \SQ{2}  \\[2pt]
    \SQ{10} & \SQ{7}  & \SQ{11} & \SQ{8}  \\[2pt]
    \SQ{16} & \SQ{13} & \SQ{17} & \SQ{14} \\[2pt]
    \SQ{22} & \SQ{19} & \SQ{23} & \SQ{20} \\[3pt]
    \footnotesize Sparse Depth &
    \footnotesize Output &
    \footnotesize Sparse Depth &
    \footnotesize Output
  \end{tabular}

  \caption{Qualitative analysis of \textbf{StarryGazer} with varying sparse point counts. Rows correspond to $n{=}50$, $200$, $500$, and $2000$ points.}
  \label{fig:sparsity_qual}
  \vspace{-2mm}
\end{figure}
\subsection{Change of Performance according to the Number of Sparse Depth Points}

In Table~\ref{tab:sparsity_comparison}, we compare our method with KBNet~[\textcolor{blue}{48}] according to the number of points in input sparse depth maps. Our method shows better MAE and RMSE values for all tested configurations (50, 200, 500, and 2000 points). This validates the robustness of our approach in handling different levels of sparsity. 
Moreover, the tendency for a larger performance gap when the number of points gets smaller shows that our model is capable of producing highly accurate estimation results even with a small amount of given information.
We present a qualitative analysis of our method in varying levels of input sparsity in Figure~\ref{fig:sparsity_qual}. It illustrates how the quality of depth completion improves as the number of input points~($n$) increases. Specifically, at n = 50, the depth map successfully captures most of the structure of the scene, showing the model's robustness even with a very sparse input despite the slight loss of some finer details. As the number of input points increases, from $n = 200$ to $n = 500$, the depth maps begin to more accurately represent the scene. The most accurate estimation is observed when n = 2000, where the depth map is considerably more detailed. This progressive enhancement underscores the importance of input density in achieving precise depth estimation, as higher point counts provide richer spatial information for the model to leverage.
\begin{table}[h!]
    \centering
    \caption{Ablations on filling the gaps on NYU Depth V2.}
    \begin{tabular}{@{}c|ccccc@{}}
    \toprule
    Method  & MAE $\downarrow$ & RMSE $\downarrow$ & \\ \midrule
    Masking out the gaps & 0.141 & 0.232 \\
    Ours & \textbf{0.100} & \textbf{0.171}  \\
    \bottomrule
    \end{tabular}
    \label{tab:fill_gap}
\end{table}
\subsection{Effect on Filling the Gaps after Rescaling}

In our training pipeline (Stage 1), gaps with no depth information are produced in synthetic depth maps because of the missing data in the segmentation masks; due to the segment sensitivity of the segmentation network, there are regions that do not belong to any segment, and such regions are transformed into gaps after the rescaling process. While just masking out these regions when calculating the loss may be a valid alternative, it can lose valuable spatial information that could contribute to the overall learning process. Moreover, masking can make the training more complicated by introducing discontinuities inside the depth map.
As described in Section 3.2, we conduct a gap-filling process to ensure continuity of the synthetic depth map. The gap-filling process can be described as follows: 
\begin{enumerate}
    \item Identify non-zero elements to determine regions with depth data.
    \item Compute the average value within the non-zero neighborhood of each zero-valued pixel by applying a $5\times5$ convolution operation, effectively smoothing over gaps.
    \item Update only the zero-valued elements in the depth image, preserving original depth values where they exist.
\end{enumerate}
This process enhances the continuity and quality of the synthetic depth images, which is crucial for effectively training the depth completion model.
We conduct experiments to compare the effect of applying an average filter to fill in the gaps versus masking the gaps from the loss for training. The results are presented in Table~\ref{tab:fill_gap}.
By utilizing the average filter to fill gaps or holes, the process yields a more consistent and continuous training dataset, which in turn enhances the model's overall performance.

\begin{table}[h!]
    \centering
    \caption{Performance comparison with different Monocular Depth Estimation (MDE) models on the NYU Depth V2 dataset.}
    \label{tab:ablation_mde}

    \begin{tabular}{c|cc}
        \toprule
        MDE Model & MAE $\downarrow$ & RMSE $\downarrow$ \\
        \midrule
        ZoeDepth~[\textcolor{blue}{15}] & $0.092$ & $0.171$ \\
        Metric3D V2~[\textcolor{blue}{35}] & $0.112$ & $0.186$ \\
        UniDepth V2~[\textcolor{blue}{36}] & $0.070$ & $0.152$ \\
        DepthAnything~[\textcolor{blue}{18}] & $0.100$ & $0.171$ \\
        DepthAnything V2~[\textcolor{blue}{37}] & $0.095$ & $0.169$ \\
        \bottomrule
    \end{tabular}
\end{table}

\subsection{Ablations on the Type of backbone MDE models}
As shown in Table~\ref{tab:ablation_mde}, we evaluate the effect of MDE models on the final performance by replacing the DepthAnything~[\textcolor{blue}{18}] model with other Monocular Depth Estimation~(MDE) models. Compared to DepthAnything, UniDepth V2~[\textcolor{blue}{36}] shows substantially better performance, reducing MAE by about 30\% and also improving RMSE. We attribute the result to the better generalizability of the model.

\begin{table}[h!]
\centering
\caption{Inference time comparison with existing depth completion methods.}
\label{tab:InferenceTime}

\begin{tabular}{c|c}
\toprule
Method & Inference time (ms) \\
\midrule
SS-S2D~[\textcolor{blue}{12}]        & 80 \\
DFuseNet~[\textcolor{blue}{13}]    & 80 \\
DDP~[\textcolor{blue}{44}]          & 80 \\
VOICED~[\textcolor{blue}{51}]        & 44 \\
AdaFrame~[\textcolor{blue}{53}] & 40 \\
SynthProj~[\textcolor{blue}{49}] & 60 \\
ScaffNet~[\textcolor{blue}{52}]    & 32 \\
KBNet~[\textcolor{blue}{50}]          & 16 \\
\midrule
Ours                        & \textbf{89} \\
\bottomrule
\end{tabular}

\end{table}

\section{Inference Time Comparison}
We measure and compare the inference time in Table~\ref{tab:InferenceTime} with a target depth map resolution of 304 by 228 pixels. While our methods require two large models (monocular depth estimation and semantic segmentation) for training, only the MDE model is used in the inference phase since we do not generate synthetic pairs. In the main experiment, we use DepthAnything with a ViT-S backbone that has the smallest number of parameters among the available configurations to mitigate the increase in inference time. Despite showing a longer inference time, we argue that our method is still meaningful, considering the improved performance and the practical applicability of our approach.

\newcommand{\chkmark}{\textcolor{blue}{\ding{51}}}
\newcommand{\crossmark}{\textcolor{red}{\ding{55}}}
\begin{table}[h!]
    \centering
\caption{Ablations on the type of segment maps used for generating synthetic dense depth maps.}
    \begin{tabular}{cc|cc}
            \toprule
            $M_{\text{seg}}$ & $I_{\text{seg}}$ & MAE $\downarrow$ & RMSE $\downarrow$ \\
            \midrule
            \crossmark & 
            \chkmark & 0.167 & 0.266 \\
            \chkmark & \chkmark & 0.124 & 0.188 \\
            \midrule
            \chkmark & \crossmark & \textbf{0.100} & \textbf{0.171} \\       
            \bottomrule
    \end{tabular}
\label{tab:ablation_seg}
\end{table}
\begin{figure}[!t]
  \centering
  \setlength{\tabcolsep}{2pt}            
  \renewcommand{\arraystretch}{0}         

  \newcommand{\figgap}{4pt}               
  \newcommand{\labpad}{2pt}               

  \newcommand{\tilew}{.238\columnwidth}
  \newcommand{\WOMDE}[1]{%
    \includegraphics[width=\tilew,page=#1]{figure_assets/woMDE.pdf}%
  }
  \newcommand{\lab}[1]{%
    \parbox[t]{\tilew}{\centering\footnotesize\vspace{\labpad}#1}%
  }

  \begin{tabular}{@{}cccc@{}}
    \WOMDE{7} & \WOMDE{1} & \WOMDE{10} & \WOMDE{4} \\[2pt]
    \WOMDE{8} & \WOMDE{2} & \WOMDE{11} & \WOMDE{5} \\[\figgap]
    \lab{RGB} & \lab{GT} & \lab{$I_{\text{seg}}$} & \lab{$M_{\text{seg}}$}
  \end{tabular}

  \caption{Qualitative comparison of depth completion using RGB-based segmentation ($I_{\text{seg}}$) and relative depth-based segmentation ($M_{\text{seg}}$) for synthetic data generation.}
  \label{fig:woMDE}
  \vspace{-2mm}
\end{figure}

\section{Alternative Approach without MDE Models}

In our approach, synthetic training pairs are typically generated by applying segmentation to relative depth maps using MDE models~[\textcolor{blue}{36, 37}]. However, MDE models may struggle in complex scenes, leading to inaccuracies.  To address the challenge, we propose an alternative approach that bypasses the need for MDE models by directly segmenting the RGB image and applying affine transformations based on grayscale values to create synthetic pairs.

We compare three strategies for generating synthetic dense depth: using segmentation from depth maps ($M_{\text{seg}}$), from RGB images ($I_{\text{seg}}$), and a mixed strategy randomly choosing between the two per iteration. As shown in Table~\ref{tab:ablation_seg}, $M_{\text{seg}}$  consistently yields the most accurate results, with lower MAE and RMSE. $I_{\text{seg}}$  performs worse due to over-segmentation in regions with uniform depth, while the mixed strategy gives intermediate results.

Qualitatively shown in Figure~\ref{fig:woMDE}, $M_{\text{seg}}$  produces smoother and more stable depth predictions, while $I_{\text{seg}}$  can better preserve fine boundaries. This flexibility allows our method to adapt depending on the availability and quality of MDE outputs. $M_{\text{seg}}$  is ideal when MDEs are reliable, while $I_{\text{seg}}$  provides a viable alternative in their absence.



